\pdfoutput=1

\documentclass[10pt, a4paper]{article}

\usepackage[]{lrec-coling2024} 

\usepackage{times}
\usepackage{multirow}
\usepackage{latexsym}

\usepackage[T1]{fontenc}

\usepackage[utf8]{inputenc}

\usepackage{microtype}
\usepackage{booktabs}
\usepackage[hang,flushmargin]{footmisc}
\usepackage{graphicx}
\DeclareGraphicsExtensions{.pdf}
\setkeys{Gin}{pagebox=artbox}

\graphicspath{{./}}

\newcommand{\bsfigure}[3][]{%
	\begin{figure}[t]
		\centering
		\includegraphics[#1]{#2}
		\caption{#3}\label{#2}%
 	 \end{figure}
}

\newcommand{\hwfigure}[3][t!]{%
	\begin{figure*}[#1]
		\centering
		\includegraphics[scale=1.0]{#2}
    		\caption{#3}\label{#2}%
  	\end{figure*}
}

\RequirePackage{type1cm}
\RequirePackage{color}
\RequirePackage{soul}
\setstcolor{blue}
\definecolor{violet}{rgb}{0.5,0.0,0.5}
\newsavebox\bscombox
\newcommand{\bscom}[3][]{%
	\sbox{\bscombox}{\fontsize{8}{9}\selectfont#1#2#3}
	\noindent
	\st{#2}{\selectfont
		\color{blue}#3\ifx\\#1\\\else{\fontsize{8}{9}\selectfont\color{violet}[#1]}\fi
	}
}

\begin{document}
\title{Modeling the Quality of Dialogical Explanations}

\name{Milad Alshomary {$^\dagger$}, Felix Lange {$^\star$},  Meisam Booshehri {$^\ddagger$},  \\
	{\bf \large Meghdut Sengupta{$^\ast$}, Philipp Cimiano{$^\ddagger$} and Henning Wachsmuth{$^\ast$}}
}

\address{$^\dagger$ Columbia University, {$^\star$}Paderborn University,  $^\ast$ Leibniz University Hannover,  $^\star$ University of Bielefeld \\
	New York, USA;  Paderborn, Germany; Hannover, Germany; Bielefeld, Germany \\
	milad.alshomary@columbia.edu\\}

\abstract{
Explanations are pervasive in our lives. Mostly, they occur in dialogical form where an {\em explainer} discusses a concept or phenomenon of interest with an {\em explainee}. Leaving the explainee with a clear understanding is not straightforward due to the knowledge gap between the two participants. Previous research looked at the interaction of explanation moves, dialogue acts, and topics in successful dialogues with expert explainers. However, daily-life explanations often fail, raising the question of what makes a dialogue successful. In this work,  we study explanation dialogues in terms of the interactions between the explainer and explainee and how they correlate with the quality of explanations in terms of a successful understanding on the explainee's side. In particular, we first construct a corpus of 399 dialogues from the Reddit forum {\em Explain Like I am Five} and annotate it for interaction flows and explanation quality. We then analyze the interaction flows, comparing them to those appearing in expert dialogues. Finally, we encode the interaction flows using two language models that can handle long inputs, and we provide empirical evidence for the effectiveness boost gained through the encoding in predicting the success of explanation dialogues.
\\ \newline \Keywords{Corpus, Explanation, Discourse Annotation,  Explainability} }

\maketitleabstract
\section{Introduction}

Explanations play a significant role in our daily life. Typically, they are realized through dialogues, where one person is an {\em explainer} while the other takes the {\em explainee} position. The explainer's primary goal is to convey information about a particular concept or phenomenon to the explainee clearly and concisely. However, ensuring that the explainee understands an explanation successfully is challenging: Effective explanations require more than just information delivery. Expert explainers usually plan an explanation strategy by choosing appropriate explanation moves, dialogue acts, and topics to ensure optimal comprehension on the explainee side \cite{wachsmuth:2022}. Additionally, explainees may actively engage in dialogues by asking clarification questions and providing feedback to ensure they understand the information correctly \cite{madumal:2019}. 

Most previous research has studied monological explanations \cite{fan:2019, situ:2021}, where an explainer provides a single-turn explanation, ignoring the role of the explainee in the interaction. However, \citet{rohlfing:2021} emphasized that both participants construct real-life explanations. While \newcite{madumal:2019} and \newcite{wachsmuth:2022} found insightful interaction patterns in the dialogues between explainers and explainees, it remains unstudied what makes such dialogues successful. In daily life, explanations may fail, depending on various factors, including the level of expertise, communication style, and prior knowledge of the explainer and the explainee. Hence, building tools to assess humans in constructing successful explanation dialogues is crucial.

\bsfigure{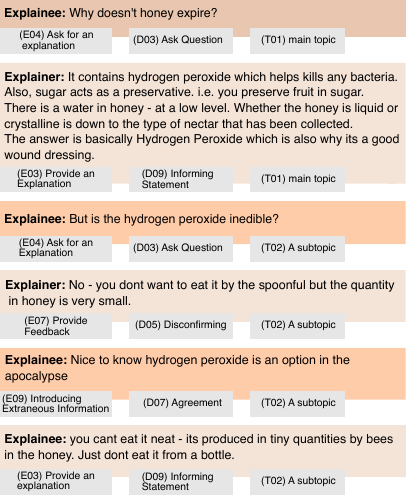}{Example explanation dialogue from the ELI5 corpus introduced in this paper, annotated for explanation moves, dialogue acts, and topics}

In this work, we take a first substantial step towards studying the quality of daily-life explanation dialogues concerning the explainee's understanding. We hypothesize that the interactions in explanation dialogues in terms of explanation moves, dialogue acts, and topics correlate with explanation success. To study this hypothesis, we construct the first corpus of daily-life explanation dialogues. We then compare this corpus to existing expert explanation dialogues \cite{wachsmuth:2022}, and we evaluate the effectiveness of pre-trained language models in predicting the quality of explanation dialogues. 

In particular, the created corpus consists of 399 daily-life explanation dialogues from the Reddit forum {\em ``Explain Like I am Five (ELI5)''}. One example dialogue is shown in Figure~\ref{exp-dlg-example.pdf}. We annotate the corpus for the explanation quality of each dialogue as well as the interaction concepts of \citet{wachsmuth:2022} (Section~\ref{sec:data}). Given the corpus, we analyze differences between daily-life and expert explanations in terms of explanation moves, dialogue acts, and topic relations. 
Matching intuition, we find that disagreement arises more often in daily life, reflecting the challenges an explainer faces while explaining a topic (Section~\ref{sec:analysis}).

To operationalize our findings, we assess whether a computational encoding of interaction flows can aid pre-trained language models in predicting the success of explanation dialogues. Specifically, we consider two popular language models on this task, namely \emph{Longformer} \cite{beltagy:2020} and \emph{hierarchical attention transformers} \cite{chalkidis:2022}. As shown in Figure~\ref{approach_new.pdf}, we augment their input with the interaction flow by prefixing each turn with its explanation move, dialogue act, and topic label. Our experiments show that adding all turn labels into the input of the \emph{hierarchical attention transformers} results in the best error reduction on the task (Section~\ref{sec:approach}). 

To summarize, the main contributions of this paper are the following: 
\begin{itemize}
	\setlength{\itemsep}{0pt}
	\item 
	A corpus of daily-life explanation dialogues annotated for interaction flow and quality
	\item 
	Insights into the differences between daily-life and expert explanation dialogues
	\item First computational approaches to the assessment of explanation dialogues.%
\end{itemize}

To foster future research, we make our corpus and all code publicly available.\footnote{data can be found in the supplementary material}

\hwfigure{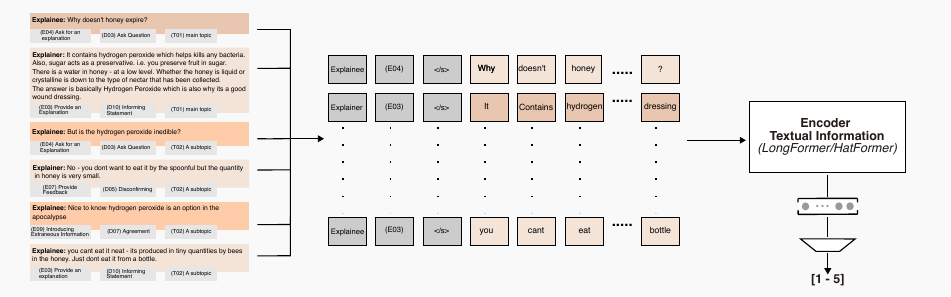}{Our approach is to augment the input of language models with tokens reflecting the interaction flow in terms of either the explanation moves, dialogue acts, topic, or all together. Here, it is shown for the case of explanation moves.}

\section{Related Work}
\label{sec:related_work}

Explanations have long been rather neglected in NLP research, but have recently gained more attention due to the increasing importance of explainable AI, XAI in short \cite{danilevsky:2020}. For XAI in general, \newcite{confalonieri:2019} discussed what make a good explanation, and \newcite{halliwell:2022} pointed out that assessing the quality of explanations is as important as it is challenging due to missing ground-truth information. The impact of the audience of an explanation was noted by \newcite{barriedoarrieta:2020}, which matches the social science view on explanations.

In particular, \newcite{miller:2019} emphasized the social aspects of explanations, arguing that explanation success depends not only on the quality of what is being said, but also on who is involved,  the social context, and what actually needs to be explained in this context. \newcite{rohlfing:2021} build on this view, clarifying that explaining in an intrinsically dialogical process in which the participants co-construct an explanation. They highlight the importance of successful communication between the explainer and explainee, which is a challenge that research needs to address adequately. 

Nevertheless, most NLP research so far focused on one-way explanations ignoring the role of the explainee. In early work \newcite{jordan:2006} analyzed the explanations of learners, whereas \newcite{fontan:2008} modeled the discourse structure of monological explanations. \newcite{jansen:2016} modeled the required explanations on exam answers, and \newcite{son:2018} looked at causality in explanations. Like us, \newcite{fan:2019} constructed a corpus of question answers from the Explain Like I am Five (ELI5) subreddit, an online community that provides simple explanations for questions asked by users. However, the instances are single question-answer pairs. \newcite{situ:2021} proposed an approach to explain machine learning models' behavior by highlighting essential parts of the input based on their contribution to the model's decision. \newcite{wiegreffe:2021} gives an overview of available datasets used in literature to model explanation in the field of XAI. Our work focuses on analyzing daily-life explanation dialogues rather than single-turn explanations.

Two related works have recently studied explanation dialogues: \newcite{madumal:2019} analyzed the transcripts of 398 explanation dialogues in terms of explainer and explainee interactions and proposed an interaction protocol to model these interactions. \newcite{wachsmuth:2022} collected a dataset of 65 explanation dialogues between an expert and explainees of different expertise levels. They proposed a taxonomy to model explanation dialogues on three dimensions, dialogue acts, explanation moves, and topics. Similar to these works, we also deal with explanation dialogues, but we aim to assess the explanation quality of the dialogues computationally.

\section{Corpus Construction}
\label{sec:data}

This section introduces our procedure to acquire our corpus from the {\em ``Explain Like I am Five (ELI-5)''} subreddit and to annotate it for our purposes.

\subsection{Explanation Dialogue Acquisition}

As mentioned in Section \ref{sec:related_work}, most existing explanation datasets target single-turn explanations and are not in a dialogical form. Therefore, we curate a new explanation dialogue corpus. Similar to \citet{fan:2019}, we use the {\em "Explain Like I am Five (ELI5)"} subreddit; however, we collect multi-turn dialogues rather than single explanations. As exemplified in Figure~\ref{exp-dlg-example.pdf}, on ELI5, a user (the \emph{explainee}) posts a question about a particular topic requesting an easy-to-understand explanation. Others (\emph{explainers}), in turn, interact with the explainee providing explanations. In some threads, these interactions turn into dialogues where the explainee elaborates their explanation need with questions or feedback, while explainers respond with clarifications. Such threads are in our focus.

For acquisition, we used the \textit{Pushshift Reddit API}.%
\footnote{\url{https://github.com/pushshift/api}}
We collected posts over a span of three years (2019--2021), creating queries to extract the top 100 threads in terms of the number of comments in each month. For each thread, we extracted explanation dialogues as follows: We identified the thread creator as the explainee. We then extracted the first-level comments that scored high up-votes since they might consist of longer discussions, and we identified their authors as explainers. For each of these comments, we searched through the nested comments to extract the alternating interactions between the explainee and the explainer. To obtain meaningful interactions, we selected only threads with a minimum of six turns for the corpus.

This process resulted in 399 explanation dialogues, covering 204 questions that we consider to be the topics of the dialogues. The dialogues have a minimum of six turns, a maximum of 40, and an average length of 8.7 turns. The corpus consists of 3457 turns with an average of 64 tokens. We call this corpus {\em ELI5-dialogues}.

\subsection{Flow and Quality Annotation}

In this work, we study what types of interactions emerge in daily-life explanation dialogues and whether certain interaction patterns correlate with a dialogue's quality in terms of a successful understanding by the explainee. Therefore, we annotate our corpus with turn-level labels reflecting the interaction flow and quality scores on the dialogue level. In the following, we first summarize definitions of the annotation scheme we reused from previous work, explain the new annotation of explanation quality, and then describe the process we followed to annotate our corpus accordingly.

\subsubsection{Interaction Flow} 
We annotate the role of each turn in an explanation dialogue following the annotation scheme of \newcite{wachsmuth:2022}. The authors proposed three aspects of interest to model explanation dialogues: explanation moves, dialogue acts, and topic relation. In the following, we describe each dimension in detail.

\paragraph{Explanation Moves}

\newcite{wachsmuth:2022} devised 10 explanation-specific moves that appear in explanation dialogues: (e$_1$) \emph{Test understanding}, identifying whether the explainee understood what has been explained (e$_2$) \emph{Test prior knowledge}, checking the explainee's level of expertise; (e$_3$) \emph{Provide explanation}, explaining any concept or a topic; (e$_4$) \emph{Request explanation}, asking for an explanation (e$_5$) \emph{Signal understanding} and (e$_6$) \emph{Signal non-understanding} to indicate that what has been said is understood or not; 
(e$_7$) \emph{Provide feedback}, responding by correcting errors or similar; (e$_8$) \emph{Provide assessment}, responding by rephrasing previous utterance or giving a hint; (e$_{9}$) \emph{Provide extra information}, providing additional information to foster a complete understanding; (e$_{10}$) \emph{Other}, representing any other move.

\paragraph{Dialogue Acts} 

As the authors, we also considered 10 dialogue acts from a standard taxonomy%
\footnote{Taxonomy of Dialogue Acts, \url{https://dit.uvt.nl}} 
to represent the communicative functions of turns: (d$_1$) \emph{Check question}, (d$_2$) \emph{What/How question}, (d$_3$) \emph{Other question}, (d$_4$) \emph{Confirming answer}, (d$_5$) \emph{Disconfirming answer}, (d$_6$) \emph{Other answer},  (d$_7$) \emph{Agreeing statement},(d$_8$) \emph{Disagreeing statement}, (d$_9$) \emph{Informing statement}, and (d$_{10}$) \emph{Other}.

\paragraph{Topic Relation} 

Capturing the relation between the turn-level and main topics can reveal different dynamics of explanation dialogues \cite{garfinkel:2009}. Therefore, we follow \newcite{wachsmuth:2022} in annotating four types of relatedness: (t$_1$) {\em Main Topic}, when the main topic is discussed; (t$_2$) {\em Suptopic}, representing a specific aspect of the main topic (e.g., {\em Music} and {\em Musical Instruments}); (t$_3$) {\em Related topic} another topic that is related to the main topic (e.g., {\em Black holes} and {\em Gravity}); and (t$_4$) {\em No/Other topic}, representing no change in the topic from previous turns.

%

\subsubsection{Explanation Quality}
Several works have explored different quality dimensions of explanations, including trustworthiness and informativeness \cite{barriedoarrieta:2020}. However, in order to avoid imposing assumptions about what makes an explanation dialogue successful, we follow a more straightforward approach to give a holistic score for each dialogue on a 5-point Likert scale, reflecting how satisfied the explainee is with the provided explanation. A score of 1 implies a fully failed explanation (no understanding visible), whereas a score of 5 means that the explanation was fully satisfactory (understanding clearly visible). Scores in between reflect different degrees of success in between.

\subsubsection{Annotation Process}
Specifically, we set up the 399 dialogues using the label-studio annotation tool \cite{maxim:2020} and recruited annotators on the UpWork platform.%
\footnote{Upwork, https://www.upwork.com}
We hired three content editors who are native English speakers and had more than 90\% job success on the platform. The annotation task was to read the whole dialogue and perform two-level annotations on the turn and dialogue levels. The annotators were asked to choose an explanation move, a dialogue act, and a topic for each turn and score the dialogue. Annotation guidelines are provided in the supplementary materials. In terms of Fleiss' $\kappa$, the annotators had an agreement of 0.73 for explanation moves, 0.49 for acts, and 0.43 for topic relation. While these values partly reflect moderate agreement, they are still notably better than those reported by \citet{wachsmuth:2022}. For the quality scores, the agreement was 0.61 in terms of ordinal Krippendorff's~$\alpha$. We consolidated the annotations using MACE \cite{hovy:2013} and split the corpus per topic question into 154 topics for training and 50 for testing. Table \ref{table-annotations} shows the frequency of each of the annotated labels in the training and test splits.
\section{Corpus Analysis}
\label{sec:analysis}

\begin{figure*}[t]
	\includegraphics{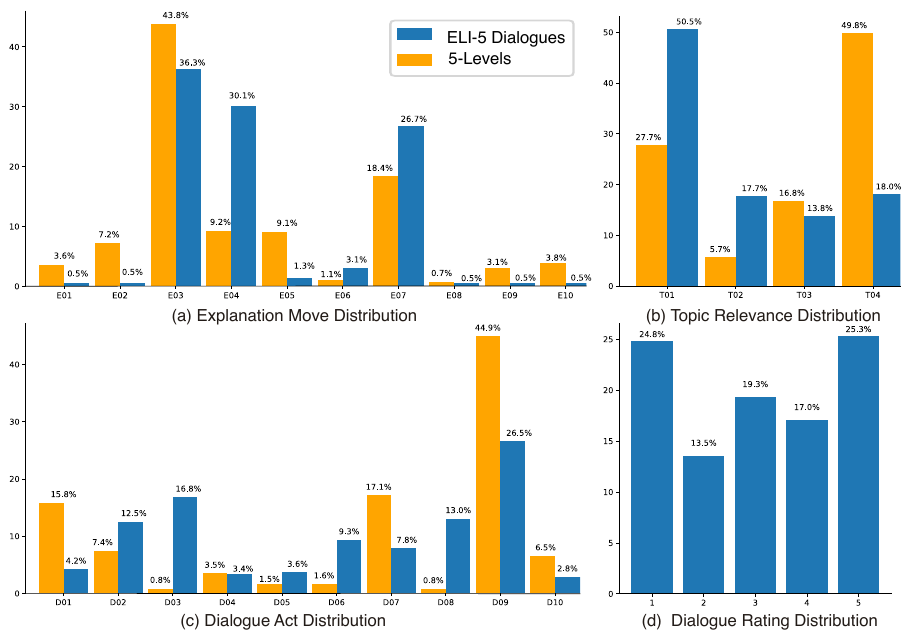}
	\caption{Explanation moves, dialogue acts, and topics distributions in our corpus and the 5-Levels the expert dialogues corpus of \citet{wachsmuth:2022}}
	\label{turn-labels-dist}
\end{figure*}

\begin{table}[t!]
	\small
	\renewcommand{\arraystretch}{0.96}
	\setlength{\tabcolsep}{3pt}
	\centering
	\begin{tabular*}{\linewidth}{llrrcrr}
		\toprule
		&					& \multicolumn{2}{r}{\bf Train}	&& \multicolumn{2}{r}{\bf Test} \\
		\cmidrule(l@{4pt}r@{0pt}){3-4}		\cmidrule(l@{4pt}r@{0pt}){6-7}		
		\multicolumn{2}{l}{\bf Turn Label}		& \bf \#	& \bf  \%				&& \bf \#	& \bf  \%	\\	
		\midrule
		(t$_1$) & Main topic & 1411 & 51.7 && 336 & 46.1 \\
		(t$_2$) & Subtopic    & 517  & 19   &&  94 & 12.9\\
		(t$_3$) & Related topic & 346  & 12.7 && 130 & 17.8 \\
		(t$_4$) & No/Other topic & 454 & 16.6 && 169 & 23.2\\
		
		\midrule
		(e$_1$) & Test understanding & 12 &  0.4  &&  5 &  0.7 \\
		(e$_2$) & Test prior knowledge & 13  & 0.5  &&  4  & 0.5\\
		(e$_3$) & Provide explanation & 1012 & 37.1 && 244  & 33.5 \\
		(e$_4$) & Request explanation & 823 & 30.2  && 217  & 29.8 \\
		(e$_5$) & Signal understanding & 36 &  1.3 &&   9 &  1.2\\
		(e$_6$) & Signal non-underst. & 85  & 3.1  && 23  & 3.2\\
		(e$_7$) & Provide feedback & 711 &  26.1 && 213 & 29.2 \\
		(e$_8$) & Provide assessment &14 &  0.5  &&  4 &  0.5\\
		(e$_9$) & Provide extra. Inf. & 13  & 0.5  &&  3 &  0.4\\
		(e$_{10}$) & Other & 9  & 0.3   && 7  & 1\\
		
		\midrule
		(d$_1$) & Check question & 113  & 4.1 &&  32  & 4.4\\
		(d$_2$) & What/How question & 349 & 12.8 &&  83 & 11.4\\
		(d$_3$) & Other question & 462 & 16.9 && 118 & 16.2\\
		(d$_4$) & Confirming answer & 87 &  3.2 &&  29  & 4\\
		(d$_5$) & Disconfirming answer & 105 &  3.8 &&  21 &  2.9\\
		(d$_6$) & Other answer & 252 &  9.2  && 70 &  9.6\\
		(d$_7$) & Agreeing statement & 192 &  7  &&   79  & 10.8\\
		(d$_8$) & Disagreeing statement & 364 & 13.3  && 86  &11.8\\
		(d$_9$) & Informing statement & 733 & 26.9 && 184 & 25.2 \\
		(d$_{10}$) & Other  & 71  & 2.6 &&  27  & 3.7\\
		

		\bottomrule
	\end{tabular*}
	\caption{Turn label distribution in our corpus for the training and testing splits.} 
	\label{table-annotations}
\end{table}

In the following, we give insights into the nature of daily-life explanations by contrasting the corpus dialogues with the expert dialogues from the {\em 5-Levels corpus} \cite{wachsmuth:2022}.

\paragraph{Turn Labels}

Figure \ref{turn-labels-dist}a illustrates the distribution of explanation moves in the two corpora. Similar to the expert dialogues, the most frequent three explanation moves in our corpus are \emph{provide explanation} (e$_3$), \emph{request explanation} (e$_4$), and \emph{provide feedback} (e$_7$), appearing in 36.3\%, 30.1\%, and 26.7\% of the turns respectively. This is expected since all dialogues in our corpus start with a question that requests an explanation. In contrast, we see that our corpus contains a higher percentage of the move \emph{signal non-understanding} (e$_6$, 3.1\%), indicating the difficulty in achieving successful explanation dialogues. Moreover, unlike the expert dialogues, the daily-life dialogues contain few turns in which the explainer \emph{tests the prior knowledge} of the explainee (e$_2$, 0.5\% compared to 7.2\%), or \emph{test their understanding} (e$_1$, 0.5\% compared to 3.6\%). 

As highlighted in Figure \ref{turn-labels-dist}c, the most frequent dialogue act is \emph{informing statement} (d$_{09}$), as in the 5-Levels corpus. However, we can see in our corpus fewer \emph{check questions}  (d$_1$) and \emph{agreeing statements} (d$_7$) compared to the expert dialogues, but more of \emph{disagreeing statements} (d$_8$). We attribute this to the fact that the explainer puts little effort into checking the understanding of the explainee --- a move can be achieved by asking check questions. Besides, the controlled setup of the expert dialogues of \newcite{wachsmuth:2022} results in much agreement between the explainer and the explainee, while in daily-life dialogues, disagreement is more prominent. 

As for the topic distribution Figure \ref{turn-labels-dist}b, in our corpus, the discussed topic in each turn is primarily the \emph{main topic} (50.5\% of the turns) followed by the subtopic (17.7\%) and others (18.0\%). In contrast, related topics are much more discussed in expert dialogues than subtopics. 

\begin{table*}[t]%
	\centering
	\small
	\setlength{\tabcolsep}{1.5pt}%
	\renewcommand{\arraystretch}{1}
	\begin{tabular}{lrrrrrr}
		\toprule
		& & \multicolumn{5}{c}{\bf Score Distribution} \\
		\cmidrule(lr){3-7}
		{\bf Explanation Moves} & Freq. & 1 & 2 & 3 & 4 & 5 \\
		\midrule
		(E03) Provide Exp. & 1256 & 22\% & 15\% & 25\% & 17\% & 21\% \\
		(E04) Ask Exp. & 1040 & 25\% & 15\% & 22\% & 15\% & 23\% \\
		(E07) Prov. Feedback & 924 & \bf 40\% & 11\% & 13\% & 14\% & 22\% \\
		(E06) Sig. Non-Under. & 108 & \bf 53\% & 14\% & 14\% & 12\% & 7\% \\
		(E05) Sig. Under. & 45 & 20\% & 13\% & 22\% & 13\% & \bf 31\% \\
		(E08) Provide Assess. & 18 & \bf 72\% & 0\% & 22\% & 6\% & 0\% \\
		(E02) Test prior know. & 17 & \bf 59\% & 18\% & 18\% & 0\% & 6\%  \\
		(E01) Test Underst. & 17 & \bf 53\% & 24\% & 6\% & 12\% & 6\% \\
		(E10) Other & 16 & 31\% & 19\% & 31\% & 19\% & 0\% \\
		(E09) Extra. Info. & 16 & \bf 62\% & 25\% & 6\% & 0\% & 6\% \\
		\bottomrule
	\end{tabular}
	\hspace{5pt}
	\begin{tabular}{lrrrrrr}
		\toprule
		& & \multicolumn{5}{c}{\bf Score Distribution} \\
		\cmidrule(lr){3-7}
		{\bf Dialogue Acts} & Freq. & 1 & 2 & 3 & 4 & 5 \\
		\midrule
		(D09) Info. Statement & 917 & 19\% & 12\% & 23\% & 18\% & \bf 28\% \\
		(D03) Question & 580 & 24\% & 13\% & 23\% & 15\% & 25\%  \\
		(D08) Disagreement & 450 & \bf 59\% & 14\% & 18\% & 6\% & 2\%  \\
		(D02) What/how Ques. & 432 & \bf 30\% & 17\% & 20\% & 15\% & 18\% \\
		(D06) Answer & 322 & 31\% & 18\% & 11\% & 18\% & 22\%  \\
		(D07) Agreement & 271 & 17\% & 7\% & 18\% & \bf 22\% & \bf 35\%  \\
		(D01) Check Question & 145 & 32\% & 18\% & 21\% & 14\% & 15\%  \\
		(D05) Disconfirm. & 126 & \bf 39\% & 16\% & 25\% & 11\% & 10\%  \\
		(D04) Confirm. & 116 & 28\% & 13\% & 16\% & 17\% & 25\%  \\
		(D10) Other & 98 & 37\% & 13\% & 15\% & 15\% & 19\% \\
		\bottomrule
	\end{tabular}
	\caption{The frequency of explanation moves (left) and dialogue acts (right) in our dataset broken  into each of the explanation quality levels [1-5]. Highlighted in bold values that distinguish the presence of these moves in high quality dialogues compared to low quality ones.}
	\label{table-exp-move-dlg-act-dist}
\end{table*}

\begin{table*}[t]%
	\centering
	\small
	\setlength{\tabcolsep}{3.5pt}%
	\renewcommand{\arraystretch}{1}
	\begin{tabular}{llrrrrrr}
		\toprule
		&  & & \multicolumn{5}{c}{\bf Score Distribution} \\
		\cmidrule(lr){4-8}
		& {\bf Dialogue Act Flow} & Freq. & 1 & 2 & 3 & 4 & 5 \\
		\midrule
		1& Ask, Inform, Ask, Inform, Ask, Inform & 14 & 7.00\% & 0\% & 7.00\% & \bf 36.00\% & \bf 50.00\% \\
		2& Ask, Inform, Ask, Inform, Ask, Inform, Ask, Inform & 6 & 0\% & \bf 33.00\% & \bf 67.00\% & 0\% & 0\% \\
		3&Ask, Inform, Ask, Inform, Ask, Inform, Agree & 5 & 20.00\% & 20.00\% & 0\% & \bf 40.00\% & 20.00\% \\
		4&Ask, Inform, Inform, Inform, Disagree, Inform, Disagree & 2 & \bf 100.00\% & 0\% & 0\% & 0\% & 0\% \\
		5&Ask, Inform, Agree, Inform, Answer, Inform, Answer & 2 & 0\% & 0\% & 0\% & \bf 100.00\% & 0\% \\
		\bottomrule
	\end{tabular}
	\caption{The most frequent dialogue act flows in our dataset broken into their frequency in each of the explanation quality levels [1-5]. Highlighted in bold values that distinguish the presence of these flows in high quality dialogues compared to low quality ones.}
	\label{table-dlg-act-flows}
\end{table*}

\paragraph{Explanation Success}

Figure \ref{turn-labels-dist}d presents the quality score distribution of the dialogues. Our data is rather balanced, with scores of 1 (24.8\%) and 5 (25.3\%) being the most frequent. Moreover, we analyze the correlation between turn labels (explanation moves, dialogue acts, and topic relatedness) and the quality scores. Table \ref{table-exp-move-dlg-act-dist} shows the frequency distribution of each turn label broken down by the quality scores. Regarding dialogue acts, labels such as {\em disagreement statement}, {\em disconfirming answer}, and {\em what/how questions} correlate more with low-quality dialogues, while informing and agreement statements correlate more with high-quality dialogues. Unexpectedly, looking at the explanation moves, we can see that {\em testing prior knowledge}, {\em providing assessment}, and {\em testing understanding} appear most in low-quality dialogues. This could be because explainers only use these moves after failing to provide a good explanation. However, as expected, moves like signaling understanding and non-understanding correlate with high and low-quality dialogue, respectively. We further look at different turn-label sequences in our dialogues and their frequencies concerning different quality levels. In terms of dialogue acts, as shown in Table \ref{table-dlg-act-flows}, successful dialogues are those of three rounds of asking questions and providing informing statements (flows \#1 and \#3) and ending with an agreeing statement, while longer interactions indicate lower quality, especially if ended with disagreeing statement (flows \#2 and \#4).
\section{Automatic Assessment of Explanation Dialogue Quality}
\label{sec:approach}

This section presents our study of the automatic assessment of explanation dialogue quality. We investigate whether augmenting the input of language models with interaction flow (encoded via special tokens) can boost their effectiveness.

\begin{table*}[t]%
	\centering
	\small
	\setlength{\tabcolsep}{2.5pt}%
	\renewcommand{\arraystretch}{1}
	\begin{tabular}{lllrrrrrrrrrrr}
		\toprule
		& & \bf Training & \multicolumn{3}{c}{\bf Explanation Moves} & & \multicolumn{3}{c}{\bf Dialogue Acts} & & \multicolumn{3}{c}{\bf Topics} \\
		\cmidrule(lr){4-6} 		\cmidrule(lr){8-10} 		\cmidrule(lr){11-14}
		\# & \bf  Model 	& \bf Data & \bf ELI-5  & \bf 5-Levels & \bf Overall && \bf ELI-5  & \bf 5-Levels & \bf Overall && \bf ELI-5  & \bf 5-Levels & \bf Overall \\
		\midrule

		1  & BERT 							 & ELI-5 	   &  0.30 & 0.25 & 0.27         && 0.34 &   0.32 & 0.38 &&  0.35 & 0.41 & 0.41  \\
		2  &										& 5-Levels			&  0.16 &  \bf 0.39 & 0.32 &&  0.14 & 0.44 & 0.30 && 0.22 & 0.47 & 0.40 \\
	   3  &										    & Both					&  0.29 & 0.38 & \bf 0.39 && 0.37 & 0.46 & 0.47 && 0.35 & 0.48 & 0.46  \\
		\addlinespace
		
		4 & BERT-Seq 	  & ELI-5  &  0.33 &   0.21 & 0.23 &&  0.35 &   0.30 &  0.36 &&  0.33 & 0.36 & 0.37  \\
		5 &								& 5-Levels 		&  0.16 &   0.38 & 0.31 &&  0.13 &   0.43 & 0.30 &&   0.32 & 0.48 & 0.44\\
		6 &								& Both 				&   0.36 &   0.37 & 0.37 &&  0.37 &   0.46 & 0.47 &&  0.35 &   0.49 & 0.47 \\

		\addlinespace
				
		7  & RoBERTa	 & ELI-5  &   0.35 &   0.21 & 0.26 && 0.39 &   0.28 & 0.39 &&  0.38 &   0.40 &  0.42 \\
		8 &							  & 5-Levels 	 &  0.18   &  \bf 0.39 & 0.33 &&  0.16 &   0.44 & 0.32 &&  0.29 &   \bf 0.54 & 0.44 \\
		9 &							 & Both              &   0.35 &  0.35  & \bf 0.39 && 0.39 & \bf 0.48 & 0.48 && \bf 0.40 & 0.53 & \bf 0.50 \\
		\addlinespace

		10 &RoBERTa-Seq  &	ELI-5  & \bf 0.39 &  0.20 &  0.24 &&  0.38 &  0.27 & 0.36 &&  0.34 &   0.31 & 0.36 \\
		11 &								 &  5-Levels      &  0.17 &  0.34 &  0.27 &&  0.16 &   0.43 & 0.31 &&  0.29 &   0.53 & 0.42 \\
		12 &								&  Both 			  & 0.34 & 0.38 & 0.38 && \bf 0.40 & 0.47 & \bf 0.49 && 0.35 & \bf 0.54 & 0.49 \\
																 
		\bottomrule
	\end{tabular}
	\caption{The macro F$_1$-score of the four evaluated models on the turn-level prediction of explanation moves, dialogue acts, and topics in 5-fold cross validation. The best score overall for each dataset is highlighted in {\bf bold}.}
	\label{table:lable-prediction-results}
\end{table*}

\subsection{Experiment Setup}

\paragraph{Task Definition} 

Give an explanation dialogue of $n$ turns between an explainer and explainee, $d=[\tau_1, \tau_2, \cdots, \tau_n]$, the task is to predict a score $S \in [1 \cdots 5]$ reflecting the dialogue quality (in our data, $n \geq 6$). Each turn $\tau_i$ is composed of a sequence of $m$ tokens, $\tau_i = (w_1, w_2, \cdots w_m)$, and has a set of three labels; the explanation act {$e_i$}, the dialogue act $d_i$, and the topic label $t_i$. 

\paragraph{Models and Baselines} 

We evaluate two recent pre-trained language models that allow processing long sequences of texts, the {\em LongFormer} \cite{beltagy:2020} and the {\em Hierarchical Attention Transformer (HAT) } \cite{nawrot:2022}. To model the interactive nature of dialogues as shown in Figure~\ref{approach_new.pdf}, we add prefix tokens for each turn $\tau_i$, representing the speaker's role (\emph{explainer} or \emph{explainee}) and the turn's interactive role represented as $e_i$, $d_i$, or $t_i$ or all together. We then concatenate all the turns' tokens as a single sequence representing the final input to the models. We compare the effectiveness of the two language models for every setting with and without the different turn labels. We also consider the average baseline that always predicts the average score from the training set. 

\paragraph{Measures} For evaluation, we compute the root mean squared error and the mean absolute error.

\paragraph{Predicting Turn Labels}

In practice, turn labels are not automatically available at inference time but must be predicted. Therefore, we also study the task of turn label prediction and test the performance of the trained quality models when the input contains ground-truth labels versus predicted labels. Since predicting turn labels is not the main focus of this paper, we retrain models available in previous work and focus on studying the generalization of these models across domains.

In particular, we experiment with the models \emph{BERT} \cite{devlin:2019} and \emph{BERT-seq} of \citet{wachsmuth:2022}. The former simply uses BERT to predict the label from the text of a single turn, while the latter utilizes a CRF layer to model dependencies between the turn labels. Moreover, we also tested RoBERTa \cite{liu:2019} as a backbone model, resulting in another two models, \emph{RoBERTa} and \emph{RoBERTa-seq}. We perform 5-fold cross-validation for each of the four models on each of the three data sources: \emph{ELI-5}, \emph{5-Levels}, and \emph{overall}. This results in 12 models that we report their results in terms of average F$1$-score across all labels of the respective task.

\subsection{Results}
\label{sec:evaluation}

\paragraph{Predicting Turn Labels}

Table \ref{table:lable-prediction-results} presents the results of predicting explanation moves, dialogue acts, and topics, broken into the corresponding performance on the two datasets and overall. In all cases, models trained on both datasets performed best ({\bf Overall} column), indicating the benefit of collecting heterogeneous datasets that cover multiple domains for the task. As for domain generalization across the two datasets (training on one dataset and evaluating on the other), BERT model generalized best from the ELI-5 dialogues to the 5-Levels dataset in all cases. For example, when predicting dialogue acts, among all models trained on ELI-5 Dialogues and evaluated on the 5-Levels dataset (\#1, \#4,  \#7, and \#10 rows), it achieves the best F$1$-score of 0.32. Moreover, we can also notice that all models were better in generalizing from the ELI-5 Dialogues dataset to the 5-Levels dataset compared to the other way around (comparing \#1, \#4, \#7, and \#10 rows to \#2, \#5, \#8, and \#11 respectively). These results are not comparable to the results of \citet{wachsmuth:2022} because we perform five-fold validation in our experiments compared to their 13-fold validation.

On the {\em ELI-5 Dialogues} dataset, {\em Roberta-seq} achieved the best results in predicting the explanation moves (when trained on the same dataset) and dialogue acts (when trained on both datasets), resulting in an F$1$-score of 0.39 and 0.40 respectively. In predicting topic relation, {\em RoBERTa} achieved the highest F$_1$-score of 0.4 when trained on both datasets. Therefore, we use these models to predict the turn labels in the following experiment. 

\begin{table}[t]%
	\centering
	\small
	\setlength{\tabcolsep}{5pt}%
	\renewcommand{\arraystretch}{1}
	\begin{tabular}{lrrrr}
		\toprule
		& \multicolumn{2}{c}{\bf Ground Truth} & \multicolumn{2}{c}{\bf Predictions} \\
		\cmidrule(lr){2-3} \cmidrule(lr){4-5}
		\bf  Approach	& \bf RMSE & \bf MAE & \bf RMSE & \bf MAE\\
		\midrule
		Average Baseline  						& 1.60 &  1.42 & 1.60 &  1.42 \\
		\addlinespace
		HatFormer 			 			   		        & 1.42        & 1.17  			& 1.42 		& 1.17\\
		\hspace{4pt} w/ Dialogue Act  &  \bf 1.29 &  \bf $^\star$1.05 	& 1.31 		& 1.09 \\
		\hspace{4pt} w/ Expl. Move      & 1.41        &  1.21  			& 1.43 		& 1.22\\
		\hspace{4pt} w/ Topic 				  & 1.41        &  1.20 		 & 1.41 	  & 1.20 \\
		\hspace{4pt} w/ ALL 				   & 1.30        & \bf 1.05    & \bf 1.28 & \bf 1.05 \\

		\addlinespace
		LongFormer                  			       & 1.34 		  & 1.13  		   & 1.34    & 1.13 \\
		\hspace{4pt} w/ Dialogue Act   &  \bf 1.31 	& \bf 1.05 	 & \bf 1.32 	& \bf 1.06 \\
		\hspace{4pt} w/ Expl. Move      &  \bf 1.31 	  & \bf 1.05   & \bf 1.32 	  & 1.09 \\
		\hspace{4pt} w/ Topic  			   	  & 1.35 	     & 1.15 		  & 1.34 	 & 1.14 \\
		\hspace{4pt} w/ ALL 				    & 1.32 		   & 1.08 		   & 1.34 	  & 1.10\\

		\bottomrule
	\end{tabular}
\caption{Explanation quality results: Root mean squared error (RMSE) and mean absolute error (MAE) of the two models with different augmented inputs, and of the average baseline for two scenarios: In \emph{ground truth}, quality is prediced based on ground-truth turn labels. In predictions, the turn labels are predicted with the developed methods. Highlighted in {\bf bold} are best scores. $^\star$ indicates statistical significance with 95\% confidence.}
\label{table:quality-pred-results}
\end{table}

\paragraph{Predicting Explanation Quality }

To obtain reliable results for each of the evaluated models (LongFormer and HatFormer with different augmented inputs), we performed 10-fold cross-validation on the training split and evaluated an ensemble of the ten trained models on the test split. We started fine-tuning from the {\em allenai/longformer-base-4096} and {\em kiddothe2b/adhoc-hierarchical-transformer-base-4096} checkpoints and selected the best checkpoint after training for 20 epochs.

Table~\ref{table:quality-pred-results} shows the quality assessment results for the baseline and the two models with different augmented inputs, given either ground truth or predicted turn labels. Compared to the average baseline, the \emph{plain} models reduce the RMSE by 0.18 and 0.26, respectively (1.60 as opposed to 1.42 and 1.34), indicating the applicability of modeling this task automatically. Encoding ground-truth interaction flows in terms of dialogue acts resulted in a further reduction of 0.13 and 0.03 RMSE for the HatFormer and Longformer, respectively. When evaluating the models on predicted turn labels, the turn label prediction errors propagate to the effectiveness in predicting quality scores. Nevertheless, we are still able to maintain error reduction better than the baseline. In the case of the HatFormer, the lowest RMSE and MAE are 1.28 and 1.05, resulting from encoding all turn labels into the input. For the LongFormer, encoding only the dialogue act into the input maintains the best results. In practice, Using the HatFormer with all turn labels encoded in the input (HatFormer w/ALL) gives the best error reduction on the task with an RMSE of 1.28 and MAE of 1.05.



\begin{figure}
	\includegraphics{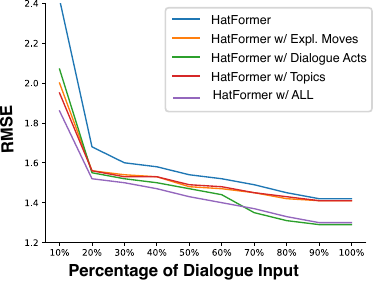}
	\caption{The root mean squared error (RSME) of all models for early predictions of explanation quality, that is, when the input to the models is only a defined initial percentage (10\%, $\ldots$, 100\%) of the full explanation dialogue.}
	\label{early_prediction_graphs}
\end{figure}

\paragraph{Early Prediction} 
As a follow-up analysis, we study the automatic quality prediction at an early stage (first few turns). From a practical viewpoint, this could be used to give the explainer insights into how the dialogue might end up so they can adjust their strategies accordingly and successfully deliver an explanation. Therefore, we evaluate the effectiveness of the HatFormer variations in terms of RMSE, when the input is only the first $k$ turns that form 10\%, 20\%, $\cdots$, 100\% of the entire dialogue (rounded). Figure~\ref{early_prediction_graphs} illustrates the results. Expectedly, the RMSE of all the models decreases with increasing the dialogue portion taken as an input. Encoding turn labels into the HatFormer results in a reduction of the RMSE for all tested dialogue proportions. Overall, we observe that at the 70\% mark, the HatFormer w/ Dialogue Acts and HatFormer w/ ALL already achieved good results.

\paragraph{Qualitative Analysis}
Upon close inspection of the explanation dialogues in our dataset, we observed that a successful explanation dialogue typically contains a pattern of requesting and providing explanations and feedback with agreeing statements. However, a low-quality dialogue consists of a sequence of providing explanations without feedback along with disagreeing statements.  
In predicting quality scores of dialogues, we found that focusing only on the dialogue content is insufficient to infer its quality. Explanation-level interactions can give the model better signals that help predict accurate scores.  Since LREC-COLING does not allow appendices at submission time, we refrain from showing example dialogues here, but we will add it upon acceptance.

\section{Conclusion}

We studied real-life explanation dialogues and how to assess their success. To this end, we constructed a dataset of real-life explanation dialogues from the {\em Explain Like I am Five} Subreddit. We annotated it according to the explanation taxonomy of \citet{wachsmuth:2022} and rated the quality of these dialogues in terms of the explainee's understanding. Our analysis provides insights into the difference between these dialogues and expert explanation dialogues. We then assessed the performance of pre-trained language models in predicting the quality of explanation dialogues and found that encoding specific interaction flows into their input boosts effectiveness.

In quantifying the explanation dialogue quality, we relied on the annotators' intuition of guessing the explainee's understanding. Although this might be a good proxy, it does not sometimes reflect the real understanding of the explainee. Moreover, in encoding dialogue interactions, better methods could be explored in the future, such as  LSTMs, Transformer-based models, or via prompting large language models LLMs.

\section*{Acknowledgments}

This work has been supported by the Deutsche For\-schungs\-ge\-mein\-schaft (DFG, German Research Foundation) under project number TRR 318/1 2021 -- 438445824. We thank the anonymous freelancers on Upwork for their annotations of our corpus.

\section{Ethics}
While constructing our corpus, we did not crawl any personal information that revealed the authors' identity. We ensured that the workers who annotated our corpus got paid more than the minimum wage in the U.S., namely 250\$ for a workload of 20 hours. 

As for the ethical implications of our work, we would like to emphasize that our research only aims to give insights into the nature of interactions in explanation dialogues. These insights and the constructed corpus can enable research in the field of explainable AI to design systems to explain model decisions to humans optimally.

\nocite{*}
\section{Bibliographical References}\label{sec:reference}
\bibliography{lrec-coling24-explanation-quality-assessment-lit}

\begin{thebibliography}{52}
\expandafter\ifx\csname natexlab\endcsname\relax\def\natexlab#1{#1}\fi

\bibitem[{Al~Khatib et~al.(2020)Al~Khatib, V{\"o}lske, Syed, Kolyada, and
  Stein}]{alkhatib:2020}
Khalid Al~Khatib, Michael V{\"o}lske, Shahbaz Syed, Nikolay Kolyada, and Benno
  Stein. 2020.
\newblock \href {https://doi.org/10.18653/v1/2020.acl-main.632} {Exploiting
  personal characteristics of debaters for predicting persuasiveness}.
\newblock In \emph{Proceedings of the 58th Annual Meeting of the Association
  for Computational Linguistics}, pages 7067--7072, Online. Association for
  Computational Linguistics.

\bibitem[{Al~Khatib et~al.(2016)Al~Khatib, Wachsmuth, Kiesel, Hagen, and
  Stein}]{alkhatib:2016b}
Khalid Al~Khatib, Henning Wachsmuth, Johannes Kiesel, Matthias Hagen, and Benno
  Stein. 2016.
\newblock \href {http://www.aclweb.org/anthology/C16-1324} {A news editorial
  corpus for mining argumentation strategies}.
\newblock In \emph{Proceedings of COLING 2016, the 26th International
  Conference on Computational Linguistics: Technical Papers}, pages 3433--3443.
  The COLING 2016 Organizing Committee.

\bibitem[{Al~Khatib et~al.(2018)Al~Khatib, Wachsmuth, Lang, Herpel, Hagen, and
  Stein}]{alkhatib:2018a}
Khalid Al~Khatib, Henning Wachsmuth, Kevin Lang, Jakob Herpel, Matthias Hagen,
  and Benno Stein. 2018.
\newblock \href {http://aclweb.org/anthology/P18-1237} {Modeling deliberative
  argumentation strategies on wikipedia}.
\newblock In \emph{Proceedings of the 56th Annual Meeting of the Association
  for Computational Linguistics (Volume 1: Long Papers)}, pages 2545--2555.
  Association for Computational Linguistics.

\bibitem[{{Barredo Arrieta} et~al.(2020){Barredo Arrieta},
  D\'{i}az-Rodr\'{i}guez, {Del Ser}, Bennetot, Tabik, Barbado, Garcia,
  Gil-Lopez, Molina, Benjamins, Chatila, and Herrera}]{barriedoarrieta:2020}
Alejandro {Barredo Arrieta}, Natalia D\'{i}az-Rodr\'{i}guez, Javier {Del Ser},
  Adrien Bennetot, Siham Tabik, Alberto Barbado, Salvador Garcia, Sergio
  Gil-Lopez, Daniel Molina, Richard Benjamins, Raja Chatila, and Francisco
  Herrera. 2020.
\newblock \href {https://doi.org/https://doi.org/10.1016/j.inffus.2019.12.012}
  {Explainable artificial intelligence {(XAI)}: {C}oncepts, taxonomies,
  opportunities and challenges toward responsible {AI}}.
\newblock \emph{Information Fusion}, 58:82--115.

\bibitem[{Beltagy et~al.(2020)Beltagy, Peters, and Cohan}]{beltagy:2020}
Iz~Beltagy, Matthew~E Peters, and Arman Cohan. 2020.
\newblock Longformer: The long-document transformer.
\newblock \emph{arXiv preprint arXiv:2004.05150}.

\bibitem[{Bourse and Saint-Dizier(2012)}]{bourse:2012}
Sarah Bourse and Patrick Saint-Dizier. 2012.
\newblock \href
  {http://www.lrec-conf.org/proceedings/lrec2012/pdf/137_Paper.pdf} {A
  repository of rules and lexical resources for discourse structure analysis:
  the case of explanation structures}.
\newblock In \emph{Proceedings of the Eighth International Conference on
  Language Resources and Evaluation ({LREC}-2012)}, pages 2778--2785, Istanbul,
  Turkey. European Languages Resources Association (ELRA).

\bibitem[{Bunt et~al.(2010)Bunt, Alexandersson, Carletta, Choe, Fang, Hasida,
  Lee, Petukhova, Popescu-Belis, Romary, Soria, and Traum}]{bunt:2010}
Harry Bunt, Jan Alexandersson, Jean Carletta, Jae-Woong Choe, Alex~Chengyu
  Fang, Koiti Hasida, Kiyong Lee, Volha Petukhova, Andrei Popescu-Belis,
  Laurent Romary, Claudia Soria, and David Traum. 2010.
\newblock \href
  {http://www.lrec-conf.org/proceedings/lrec2010/pdf/560_Paper.pdf} {Towards an
  {ISO} standard for dialogue act annotation}.
\newblock In \emph{Proceedings of the Seventh International Conference on
  Language Resources and Evaluation ({LREC}'10)}, Valletta, Malta. European
  Language Resources Association (ELRA).

\bibitem[{Caines et~al.(2020)Caines, Yannakoudakis, Edmondson, Allen,
  P{\'e}rez-Paredes, Byrne, and Buttery}]{caines:2020}
Andrew Caines, Helen Yannakoudakis, Helena Edmondson, Helen Allen, Pascual
  P{\'e}rez-Paredes, Bill Byrne, and Paula Buttery. 2020.
\newblock The teacher-student chatroom corpus.
\newblock In \emph{Proceedings of the 9th Workshop on NLP for Computer Assisted
  Language Learning}, pages 10--20.

\bibitem[{Chalkidis et~al.(2022)Chalkidis, Dai, Fergadiotis, Malakasiotis, and
  Elliott}]{chalkidis:2022}
Ilias Chalkidis, Xiang Dai, Manos Fergadiotis, Prodromos Malakasiotis, and
  Desmond Elliott. 2022.
\newblock \href {https://arxiv.org/abs/2210.05529} {An exploration of
  hierarchical attention transformers for efficient long document
  classification}.

\bibitem[{Chari et~al.(2020)Chari, Seneviratne, Gruen, Foreman, Das, and
  McGuinness}]{chari:2020}
Shruthi Chari, Oshani Seneviratne, Daniel~M Gruen, Morgan~A Foreman, Amar~K
  Das, and Deborah~L McGuinness. 2020.
\newblock Explanation ontology: a model of explanations for user-centered ai.
\newblock In \emph{The Semantic Web--ISWC 2020: 19th International Semantic Web
  Conference, Athens, Greece, November 2--6, 2020, Proceedings, Part II}, pages
  228--243. Springer.

\bibitem[{Confalonieri et~al.(2019)Confalonieri, Besold, Weyde, Creel,
  Lombrozo, Mueller, and Shafto}]{confalonieri:2019}
Roberto Confalonieri, Tarek~R. Besold, Tillman Weyde, Kathleen Creel, Tania
  Lombrozo, Shane~T. Mueller, and Patrick Shafto. 2019.
\newblock \href {https://mindmodeling.org/cogsci2019/papers/0013/index.html}
  {What makes a good explanation? {C}ognitive dimensions of explaining
  intelligent machines}.
\newblock In \emph{Proceedings of the 41th Annual Meeting of the Cognitive
  Science Society, CogSci 2019: Creativity + Cognition + Computation, Montreal,
  Canada, July 24-27, 2019}, pages 25--26.

\bibitem[{Danilevsky et~al.(2020)Danilevsky, Qian, Aharonov, Katsis, Kawas, and
  Sen}]{danilevsky:2020}
Marina Danilevsky, Kun Qian, Ranit Aharonov, Yannis Katsis, Ban Kawas, and
  Prithviraj Sen. 2020.
\newblock \href {https://aclanthology.org/2020.aacl-main.46} {A survey of the
  state of explainable {AI} for natural language processing}.
\newblock In \emph{Proceedings of the 1st Conference of the Asia-Pacific
  Chapter of the Association for Computational Linguistics and the 10th
  International Joint Conference on Natural Language Processing}, pages
  447--459, Suzhou, China. Association for Computational Linguistics.

\bibitem[{Devlin et~al.(2018)Devlin, Chang, Lee, and Toutanova}]{devlin:2018}
Jacob Devlin, Ming-Wei Chang, Kenton Lee, and Kristina Toutanova. 2018.
\newblock Bert: Pre-training of deep bidirectional transformers for language
  understanding.
\newblock \emph{arXiv preprint arXiv:1810.04805}.

\bibitem[{Devlin et~al.(2019)Devlin, Chang, Lee, and Toutanova}]{devlin:2019}
Jacob Devlin, Ming-Wei Chang, Kenton Lee, and Kristina Toutanova. 2019.
\newblock \href {https://doi.org/10.18653/v1/N19-1423} {{BERT}: {P}re-training
  of deep bidirectional transformers for language understanding}.
\newblock In \emph{Proceedings of the 2019 {Conference} of the {North}
  {American} {Chapter} of the {Association} for {Computational} {Linguistics}:
  {Human} {Language} {Technologies}, {Volume} 1 ({Long} and {Short} {Papers})},
  pages 4171--4186, Minneapolis, Minnesota. Association for Computational
  Linguistics.

\bibitem[{Dulceanu et~al.(2018)Dulceanu, Le~Dinh, Chang, Bui, Kim, Vu, and
  Kim}]{dulceanu:2018}
Andrei Dulceanu, Thang Le~Dinh, Walter Chang, Trung Bui, Doo~Soon Kim,
  Manh~Chien Vu, and Seokhwan Kim. 2018.
\newblock \href {https://www.aclweb.org/anthology/L18-1438}
  {{P}hotoshop{Q}ui{A}: {A} corpus of non-factoid questions and answers for
  why-question answering}.
\newblock In \emph{Proceedings of the Eleventh International Conference on
  Language Resources and Evaluation ({LREC} 2018)}, Miyazaki, Japan. European
  Language Resources Association (ELRA).

\bibitem[{Dzikovska et~al.(2012)Dzikovska, Nielsen, and Brew}]{dzikovska:2012}
Myroslava~O. Dzikovska, Rodney~D. Nielsen, and Chris Brew. 2012.
\newblock \href {https://www.aclweb.org/anthology/N12-1021} {Towards effective
  tutorial feedback for explanation questions: A dataset and baselines}.
\newblock In \emph{Proceedings of the 2012 Conference of the North {A}merican
  Chapter of the Association for Computational Linguistics: Human Language
  Technologies}, pages 200--210, Montr{\'e}al, Canada. Association for
  Computational Linguistics.

\bibitem[{Falk and Lapesa(2023)}]{falk:2023}
Neele Falk and Gabriella Lapesa. 2023.
\newblock \href {https://aclanthology.org/2023.findings-eacl.187} {Bridging
  argument quality and deliberative quality annotations with adapters}.
\newblock In \emph{Findings of the Association for Computational Linguistics:
  EACL 2023}, pages 2469--2488, Dubrovnik, Croatia. Association for
  Computational Linguistics.

\bibitem[{Fan et~al.(2019)Fan, Jernite, Perez, Grangier, Weston, and
  Auli}]{fan:2019}
Angela Fan, Yacine Jernite, Ethan Perez, David Grangier, Jason Weston, and
  Michael Auli. 2019.
\newblock \href {https://doi.org/10.18653/v1/P19-1346} {{ELI}5: {L}ong form
  question answering}.
\newblock In \emph{Proceedings of the 57th Annual Meeting of the Association
  for Computational Linguistics}, pages 3558--3567, Florence, Italy.
  Association for Computational Linguistics.

\bibitem[{Finke et~al.(2022)Finke, Horwath, Matzner, and Schulz}]{finke:2022}
Josefine Finke, Ilona Horwath, Tobias Matzner, and Christian Schulz. 2022.
\newblock (de)coding social practice in the field of xai: {T}owards a
  co-constructive framework of explanations and understanding between lay users
  and algorithmic systems.
\newblock In \emph{Artificial Intelligence in HCI}, pages 149--160, Cham.
  Springer International Publishing.

\bibitem[{Fontan and Saint-Dizier(2008)}]{fontan:2008}
Lionel Fontan and Patrick Saint-Dizier. 2008.
\newblock \href {https://aclanthology.org/W08-2210} {Analyzing the explanation
  structure of procedural texts: Dealing with advice and warnings}.
\newblock In \emph{Semantics in Text Processing. {STEP} 2008 Conference
  Proceedings}, pages 115--127. College Publications.

\bibitem[{Fried et~al.(2018)Fried, Andreas, and Klein}]{fried:2018}
Daniel Fried, Jacob Andreas, and Dan Klein. 2018.
\newblock \href {https://doi.org/10.18653/v1/N18-1177} {Unified pragmatic
  models for generating and following instructions}.
\newblock In \emph{Proceedings of the 2018 Conference of the North {A}merican
  Chapter of the Association for Computational Linguistics: Human Language
  Technologies, Volume 1 (Long Papers)}, pages 1951--1963, New Orleans,
  Louisiana. Association for Computational Linguistics.

\bibitem[{Garfinkel(2009)}]{garfinkel:2009}
Alan Garfinkel. 2009.
\newblock \emph{Forms of {Explanation}: {Rethinking} the {Questions} in
  {Social} {Theory}}, revised edition.
\newblock Yale University Press, New Haven \& London, New Haven; London.

\bibitem[{Gilpin et~al.(2018)Gilpin, Bau, Yuan, Bajwa, Specter, and
  Kagal}]{gilpin:2018}
Leilani~H. Gilpin, David Bau, Ben~Z. Yuan, Ayesha Bajwa, Michael Specter, and
  Lalana Kagal. 2018.
\newblock \href {http://arxiv.org/abs/1806.00069} {Explaining explanations:
  {A}n overview of interpretability of machine learning}.
\newblock ArXiv: 1806.00069.

\bibitem[{Goodman and Flaxman(2017)}]{goodman:2017}
Bryce Goodman and Seth Flaxman. 2017.
\newblock \href {https://doi.org/10.1609/aimag.v38i3.2741} {European union
  regulations on algorithmic decision-making and a ``right to explanation''}.
\newblock \emph{AI Magazine}, 38(3):50--57.

\bibitem[{Habernal et~al.(2018)Habernal, Wachsmuth, Gurevych, and
  Stein}]{habernal:2018a}
Ivan Habernal, Henning Wachsmuth, Iryna Gurevych, and Benno Stein. 2018.
\newblock \href {http://aclweb.org/anthology/N18-1175} {The argument reasoning
  comprehension task: Identification and reconstruction of implicit warrants}.
\newblock In \emph{Proceedings of the 2018 Conference of the North American
  Chapter of the Association for Computational Linguistics: Human Language
  Technologies, Volume 1 (Long Papers)}, pages 1930--1940. Association for
  Computational Linguistics.

\bibitem[{Halliwell et~al.(2022)Halliwell, Gandon, Lecue, and
  Villata}]{halliwell:2022}
Nicholas Halliwell, Fabien Gandon, Freddy Lecue, and Serena Villata. 2022.
\newblock \href {https://hal.science/hal-03591012} {{The Need for Empirical
  Evaluation of Explanation Quality}}.
\newblock In \emph{{AAAI 2022 - Workshop on Explainable Agency in Artificial
  Intelligence}}, Vancouver, Canada.

\bibitem[{Hovy et~al.(2013)Hovy, Berg-Kirkpatrick, Vaswani, and
  Hovy}]{hovy:2013}
Dirk Hovy, Taylor Berg-Kirkpatrick, Ashish Vaswani, and Eduard Hovy. 2013.
\newblock \href {https://aclanthology.org/N13-1132} {Learning whom to trust
  with {MACE}}.
\newblock In \emph{Proceedings of the 2013 Conference of the North {A}merican
  Chapter of the Association for Computational Linguistics: Human Language
  Technologies}, Atlanta, Georgia. Association for Computational Linguistics.

\bibitem[{Jansen et~al.(2016)Jansen, Balasubramanian, Surdeanu, and
  Clark}]{jansen:2016}
Peter Jansen, Niranjan Balasubramanian, Mihai Surdeanu, and Peter Clark. 2016.
\newblock \href {https://aclanthology.org/C16-1278} {What{'}s in an
  explanation? characterizing knowledge and inference requirements for
  elementary science exams}.
\newblock In \emph{Proceedings of {COLING} 2016, the 26th International
  Conference on Computational Linguistics: Technical Papers}, pages 2956--2965,
  Osaka, Japan. The COLING 2016 Organizing Committee.

\bibitem[{Jordan et~al.(2006)Jordan, Makatchev, and Pappuswamy}]{jordan:2006}
Pamela~W. Jordan, Maxim Makatchev, and Umarani Pappuswamy. 2006.
\newblock \href {https://aclanthology.org/W06-3503} {Understanding complex
  natural language explanations in tutorial applications}.
\newblock In \emph{Proceedings of the Third Workshop on Scalable Natural
  Language Understanding}, pages 17--24, New York City, New York. Association
  for Computational Linguistics.

\bibitem[{Li et~al.(2021)Li, Zhang, and Chen}]{li:2021}
Lei Li, Yongfeng Zhang, and Li~Chen. 2021.
\newblock \href {https://doi.org/10.18653/v1/2021.acl-long.383} {Personalized
  transformer for explainable recommendation}.
\newblock In \emph{Proceedings of the 59th Annual Meeting of the Association
  for Computational Linguistics and the 11th International Joint Conference on
  Natural Language Processing (Volume 1: Long Papers)}, pages 4947--4957,
  Online. Association for Computational Linguistics.

\bibitem[{Liu et~al.(2019)Liu, Ott, Goyal, Du, Joshi, Chen, Levy, Lewis,
  Zettlemoyer, and Stoyanov}]{liu:2019}
Yinhan Liu, Myle Ott, Naman Goyal, Jingfei Du, Mandar Joshi, Danqi Chen, Omer
  Levy, Mike Lewis, Luke Zettlemoyer, and Veselin Stoyanov. 2019.
\newblock Roberta: A robustly optimized bert pretraining approach.
\newblock \emph{arXiv preprint arXiv:1907.11692}.

\bibitem[{Madumal et~al.(2019)Madumal, Miller, Sonenberg, and
  Vetere}]{madumal:2019}
Prashan Madumal, Tim Miller, Liz Sonenberg, and Frank Vetere. 2019.
\newblock A grounded interaction protocol for explainable artificial
  intelligence.
\newblock In \emph{Proceedings of the 18th International Conference on
  Autonomous Agents and MultiAgent Systems}, AAMAS '19, page 1033?1041,
  Richland, SC. International Foundation for Autonomous Agents and Multiagent
  Systems.

\bibitem[{Mann and Thompson(1988)}]{mann:1988}
William~C Mann and Sandra~A Thompson. 1988.
\newblock Rhetorical structure theory: {T}oward a functional theory of text
  organization.
\newblock \emph{Text-interdisciplinary Journal for the Study of Discourse},
  8(3):243--281.

\bibitem[{Miller(2019)}]{miller:2019}
Tim Miller. 2019.
\newblock \href {https://doi.org/10.1016/j.artint.2018.07.007} {Explanation in
  artificial intelligence: {Insights} from the social sciences}.
\newblock \emph{Artificial Intelligence}, 267:1--38.

\bibitem[{Nakov et~al.(2017)Nakov, Hoogeveen, M{\`a}rquez, Moschitti, Mubarak,
  Baldwin, and Verspoor}]{nakov:2017}
Preslav Nakov, Doris Hoogeveen, Llu{\'\i}s M{\`a}rquez, Alessandro Moschitti,
  Hamdy Mubarak, Timothy Baldwin, and Karin Verspoor. 2017.
\newblock \href {https://doi.org/10.18653/v1/S17-2003} {{S}em{E}val-2017 task
  3: Community question answering}.
\newblock In \emph{Proceedings of the 11th International Workshop on Semantic
  Evaluation ({S}em{E}val-2017)}, pages 27--48, Vancouver, Canada. Association
  for Computational Linguistics.

\bibitem[{Nawrot et~al.(2022)Nawrot, Tworkowski, Tyrolski, Kaiser, Wu, Szegedy,
  and Michalewski}]{nawrot:2022}
Piotr Nawrot, Szymon Tworkowski, Micha{\l} Tyrolski, {\L}ukasz Kaiser, Yuhuai
  Wu, Christian Szegedy, and Henryk Michalewski. 2022.
\newblock Hierarchical transformers are more efficient language models.
\newblock In \emph{Findings of the Association for Computational Linguistics:
  NAACL 2022}, pages 1559--1571.

\bibitem[{Rohlfing et~al.(2021)Rohlfing, Cimiano, Scharlau, Matzner, Buhl,
  Buschmeier, Esposito, Grimminger, Hammer, H\"{a}b-Umbach, Horwath,
  H\"{u}llermeier, Kern, Kopp, Thommes, Ngonga~Ngomo, Schulte, Wachsmuth,
  Wagner, and Wrede}]{rohlfing:2021}
Katharina~J. Rohlfing, Philipp Cimiano, Ingrid Scharlau, Tobias Matzner,
  Heike~M. Buhl, Hendrik Buschmeier, Elena Esposito, Angela Grimminger, Barbara
  Hammer, Reinhold H\"{a}b-Umbach, Ilona Horwath, Eyke H\"{u}llermeier,
  Friederike Kern, Stefan Kopp, Kirsten Thommes, Axel-Cyrille Ngonga~Ngomo,
  Carsten Schulte, Henning Wachsmuth, Petra Wagner, and Britta Wrede. 2021.
\newblock \href {https://doi.org/10.1109/TCDS.2020.3044366} {Explanation as a
  social practice: {T}oward a conceptual framework for the social design of
  {AI} systems}.
\newblock \emph{IEEE Transactions on Cognitive and Developmental Systems},
  13(3):717--728.

\bibitem[{Situ et~al.(2021)Situ, Zukerman, Paris, Maruf, and
  Haffari}]{situ:2021}
Xuelin Situ, Ingrid Zukerman, Cecile Paris, Sameen Maruf, and Gholamreza
  Haffari. 2021.
\newblock \href {https://doi.org/10.18653/v1/2021.acl-long.415} {Learning to
  explain: Generating stable explanations fast}.
\newblock In \emph{Proceedings of the 59th Annual Meeting of the Association
  for Computational Linguistics and the 11th International Joint Conference on
  Natural Language Processing (Volume 1: Long Papers)}, pages 5340--5355,
  Online. Association for Computational Linguistics.

\bibitem[{Son et~al.(2018)Son, Bayas, and Schwartz}]{son:2018}
Youngseo Son, Nipun Bayas, and H.~Andrew Schwartz. 2018.
\newblock \href {https://www.aclweb.org/anthology/D18-1372} {Causal explanation
  analysis on social media}.
\newblock In \emph{Proceedings of the 2018 Conference on Empirical Methods in
  Natural Language Processing}, pages 3350--3359, Brussels, Belgium.
  Association for Computational Linguistics.

\bibitem[{Stolcke et~al.(2000)Stolcke, Ries, Coccaro, Shriberg, Bates,
  Jurafsky, Taylor, Martin, Van Ess-Dykema, and Meteer}]{stolcke:2000}
Andreas Stolcke, Klaus Ries, Noah Coccaro, Elizabeth Shriberg, Rebecca Bates,
  Daniel Jurafsky, Paul Taylor, Rachel Martin, Carol Van Ess-Dykema, and Marie
  Meteer. 2000.
\newblock \href {https://aclanthology.org/J00-3003} {Dialogue act modeling for
  automatic tagging and recognition of conversational speech}.
\newblock \emph{Computational Linguistics}, 26(3):339--374.

\bibitem[{Suresh et~al.(2022)Suresh, Jacobs, Harty, Perkoff, Martin, and
  Sumner}]{suresh:2022}
Abhijit Suresh, Jennifer Jacobs, Charis Harty, Margaret Perkoff, James~H.
  Martin, and Tamara Sumner. 2022.
\newblock \href {https://aclanthology.org/2022.lrec-1.497} {The {T}alk{M}oves
  dataset: K-12 mathematics lesson transcripts annotated for teacher and
  student discursive moves}.
\newblock In \emph{Proceedings of the Thirteenth Language Resources and
  Evaluation Conference}, pages 4654--4662, Marseille, France. European
  Language Resources Association.

\bibitem[{Swales(1990)}]{swales:1990}
John~M. Swales. 1990.
\newblock \emph{Genre Analysis: {E}nglish in Academic and Research Settings}.
\newblock Cambridge University Press.

\bibitem[{Tkachenko et~al.(2020-2022)Tkachenko, Malyuk, Holmanyuk, and
  Liubimov}]{maxim:2020}
Maxim Tkachenko, Mikhail Malyuk, Andrey Holmanyuk, and Nikolai Liubimov.
  2020-2022.
\newblock \href {https://github.com/heartexlabs/label-studio} {{Label Studio}:
  Data labeling software}.
\newblock Open source software available from
  https://github.com/heartexlabs/label-studio.

\bibitem[{Vander~Linden(1992)}]{vander:1992}
Keith Vander~Linden. 1992.
\newblock The expression of local rhetorical relations in instructional text.
\newblock In \emph{Proceedings of the 30th Annual Meeting of the Association
  for Computational Linguistics}, pages 318--320.

\bibitem[{Vecchi et~al.(2021)Vecchi, Falk, Jundi, and Lapesa}]{vecchi:2021}
Eva~Maria Vecchi, Neele Falk, Iman Jundi, and Gabriella Lapesa. 2021.
\newblock \href {https://doi.org/10.18653/v1/2021.acl-long.107} {Towards
  argument mining for social good: A survey}.
\newblock In \emph{Proceedings of the 59th Annual Meeting of the Association
  for Computational Linguistics and the 11th International Joint Conference on
  Natural Language Processing (Volume 1: Long Papers)}, pages 1338--1352,
  Online. Association for Computational Linguistics.

\bibitem[{Wachsmuth and Alshomary(2022)}]{wachsmuth:2022}
Henning Wachsmuth and Milad Alshomary. 2022.
\newblock \href {https://aclanthology.org/2022.coling-1.27} {{``M}ama always
  had a way of explaining things so {I} could understand{''}: {A} dialogue
  corpus for learning to construct explanations}.
\newblock In \emph{Proceedings of the 29th International Conference on
  Computational Linguistics}, pages 344--354, Gyeongju, Republic of Korea.
  International Committee on Computational Linguistics.

\bibitem[{Wachsmuth et~al.(2017)Wachsmuth, Naderi, Hou, Bilu, Prabhakaran,
  Thijm, Hirst, and Stein}]{wachsmuth:2017b}
Henning Wachsmuth, Nona Naderi, Yufang Hou, Yonatan Bilu, Vinodkumar
  Prabhakaran, Alberdingk~Tim Thijm, Graeme Hirst, and Benno Stein. 2017.
\newblock \href {http://aclweb.org/anthology/E17-1017} {Computational
  argumentation quality assessment in natural language}.
\newblock In \emph{Proceedings of the 15th Conference of the European Chapter
  of the Association for Computational Linguistics: Volume 1, Long Papers},
  pages 176--187. Association for Computational Linguistics.

\bibitem[{Wachsmuth and Stein(2017)}]{wachsmuth:2017c}
Henning Wachsmuth and Benno Stein. 2017.
\newblock \href {https://doi.org/http://doi.acm.org/10.1145/2957757} {A
  universal model for discourse-level argumentation analysis}.
\newblock \emph{Special Section of the {ACM} Transactions on Internet
  Technology: {A}rgumentation in Social Media}, 17(3):28:1--28:24.

\bibitem[{Wang et~al.(2017)Wang, Beauchamp, Shugars, and Qin}]{wang:2017}
Lu~Wang, Nick Beauchamp, Sarah Shugars, and Kechen Qin. 2017.
\newblock \href {https://doi.org/10.1162/tacl_a_00057} {Winning on the merits:
  The joint effects of content and style on debate outcomes}.
\newblock \emph{Transactions of the Association for Computational Linguistics},
  5:219--232.

\bibitem[{Wiegreffe and Marasovi{\'c}(2021)}]{wiegreffe:2021}
Sarah Wiegreffe and Ana Marasovi{\'c}. 2021.
\newblock Teach me to explain: A review of datasets for explainable natural
  language processing.
\newblock \emph{arXiv preprint arXiv:2102.12060}.

\bibitem[{Yagcioglu et~al.(2018)Yagcioglu, Erdem, Erdem, and
  Ikizler-Cinbis}]{yagcioglu:2018}
Semih Yagcioglu, Aykut Erdem, Erkut Erdem, and Nazli Ikizler-Cinbis. 2018.
\newblock \href {https://doi.org/10.18653/v1/D18-1166} {{R}ecipe{QA}: A
  challenge dataset for multimodal comprehension of cooking recipes}.
\newblock In \emph{Proceedings of the 2018 Conference on Empirical Methods in
  Natural Language Processing}, pages 1358--1368, Brussels, Belgium.
  Association for Computational Linguistics.

\bibitem[{Zhang et~al.(2012)Zhang, Webster, Uren, Varga, and
  Ciravegna}]{zhang:2012}
Ziqi Zhang, Philip Webster, Victoria Uren, Andrea Varga, and Fabio Ciravegna.
  2012.
\newblock \href
  {http://www.lrec-conf.org/proceedings/lrec2012/pdf/244_Paper.pdf}
  {Automatically extracting procedural knowledge from instructional texts using
  natural language processing}.
\newblock In \emph{Proceedings of the Eighth International Conference on
  Language Resources and Evaluation ({LREC}'12)}, pages 520--527, Istanbul,
  Turkey. European Language Resources Association (ELRA).

\end{thebibliography}
\bibliographystyle{lrec-coling2024-natbib}

\appendix
\section{Annotation Analysis}
Table \ref{table-topic-dist} shows the distribution of topic relation over different quality scores. 

\begin{table*}[h]%
	\centering
	\small
	\setlength{\tabcolsep}{3.5pt}%
	\renewcommand{\arraystretch}{1}
	\begin{tabular}{llrrrrrr}
		\toprule
		&  & & \multicolumn{5}{c}{\bf Score Distribution} \\
		\cmidrule(lr){4-8}
		& {\bf Explanation Move Flow} & Freq. & 1 & 2 & 3 & 4 & 5 \\
		\midrule
             1&Req., Explain, Req., Explain, Req., Explain,& 17 & 6\% & 0\% & 12\% & \bf  41\% & \bf 41\%\\ 
             2&Req., Explain, Req., Explain, Feedback, Feedback & 9 & 11\% & 0\% & 33\% & 0\% & \bf 56\%\\ 
             3&Req., Explain, Req., Explain, Req., Explain, Req., Explain,& 8 & 0\% & \bf 25\% & \bf 62\% & 0\% & 12\%\\ 
             4&Req., Explain, Feedback, Feedback, Feedback, Feedback & 5 & \bf 20\% & \bf 20\% & 20\% & 0\% & 40\%\\ 
             5&Req., Explain, Req., Explain, Feedback, Explain, Feedback & 4 & 0\% & \bf 50\% & 0\% & 25\% & 25\%\\ 
             6&Req., Feedback, Feedback, Feedback, Feedback, Feedback & 3 & \bf 67\% & 0\% & 0\% & 0\% & 33\%\\  
		\bottomrule
	\end{tabular}
	\caption{The most frequent explanation move flows in our dataset broken into their frequency in each of the explanation quality levels [1-5]. Highlighted in bold values that distinguish the presence of these flows in high quality dialogues compared to low quality ones.}
	\label{table-exp-move-flows}
\end{table*}

In terms of explanation moves sequences, as shown in Table \ref{table-exp-move-flows}, prominent flows that occur in high-quality explanation dialogues are those that contain two rounds of requesting and providing explanations followed by some feedback (flows \#2 and \#3). Not surprisingly, dialogues that do barely provide explanations (flow \#7 and \#8) are ranked to be of lower quality.  Regarding topic sequences, Table \ref{table-topic-flows} shows that diverging from the main topic mostly correlate with low-quality dialogues.

\begin{table*}[t]%
	\centering
	\small
	\setlength{\tabcolsep}{3.5pt}%
	\renewcommand{\arraystretch}{1}
	\begin{tabular}{llrrrrrr}
		\toprule
		&  & & \multicolumn{5}{c}{\bf Score Distribution} \\
		\cmidrule(lr){4-8}
		& {\bf Topic Relation Flow} & Freq. & 1 & 2 & 3 & 4 & 5 \\
		\midrule
			1&Main, Main, Main, Main, Main, Main & 62 & 11.00\% & 13.00\% & 13.00\% & \bf 18.00\% & \bf 45.00\%\\ 
			2&Main, Main, Main, Main, Main, Main, Main & 12 & 42.00\% & 8.00\% & 8.00\% & 42.00\% & 0\%\\ 
			3&Main, Other, Other, Other, Other, Other, Other & 9 & \bf 56.00\% & 11.00\% & 22.00\% & 11.00\% & 0\%\\ 
			4&Main, Other, Other, Other, Other, Other & 7 & \bf 100.00\% & 0\% & 0\% & 0\% & 0\%\\ 
			5&Main, Suptopic, Suptopic, Suptopic, Suptopic, Suptopic & 7 & 14.00\% & 14.00\% & 29.00\% & 29.00\% & 14.00\%\\ 
		\bottomrule
	\end{tabular}
	\caption{The most frequent topic relation flows in our dataset broken into their frequency in each of the explanation quality levels [1-5]. Highlighted in bold values that distinguish the presence of these flows in high quality dialogues compared to low quality ones.}
	\label{table-topic-flows}
\end{table*}

\section{Example Dialogues}
Table \ref{table:example-dialogues} present two example dialogues. Dialogue \#1 was annotated as a high-quality explanation dialogue while the Dialogue \#2 as low-quality.

\begin{table*}[h]%
	\small
	\setlength{\tabcolsep}{5pt}%
	\renewcommand{\arraystretch}{1}
	\begin{tabular}{lp{1.5cm}p{2cm}p{1.5cm}}
		\toprule
		{\bf  Dialogue \#1} & {\bf Rating:} 4 &&\\
		\midrule
		\multicolumn{4}{p{7cm}}{{\bf Explainee:} Why are there not many "flamboyant" heterosexual males?} \\
		\em \tiny Request Explanation & \em \tiny Other Question & \em \tiny Main Topic &\\
		\addlinespace
		
		\multicolumn{4}{p{7cm}}{{\bf Explainer:} I think a lot of the flamboyance is actually an act, albeit an unintentional one. It’s a lot about fitting in with the culture. I know a handful of “straight” guys who were “turned” by my gay friends and in a year these previously straight-acting men are the gayest of the bunch.} \\
		\em \tiny Provide Explanation & \em \tiny Informing Statement & \em \tiny Main Topic & \\
		\addlinespace
		
		\multicolumn{4}{p{7cm}}{{\bf Explainee:} Thank you for not attacking my question and seeing it for the curiosity it is. I do believe culture and fitting in does play a large role here. But I haven't run into any flamboyant heterosexual males.} \\
		\em \tiny Provide Feedback & \em \tiny Agreeing Statement & \em \tiny Main Topic & \\
		\addlinespace
		
		\multicolumn{4}{p{7cm}}{{\bf Explainer:} I guess we'd have to look at straight males that were raised by really flamboyant parents and see how they turned out.} \\
		\em \tiny Provide Feedback & \em \tiny Agreeing Statement & \em \tiny Main Topic & \\
		\addlinespace
		
		\multicolumn{4}{p{7cm}}{{\bf Explainee:} I dont know if that would be considered cruel and unusual if done purposefully. But undoubtedly there should be 2 flamboyant men that could care for a child better than at least some heterosexual couples.} \\
		\em \tiny Provide Explanation & \em \tiny Informing Statement & \em \tiny Main Topic &  \\
		\addlinespace
		
		\multicolumn{4}{p{7cm}}{{\bf Explainer:} Yea we'll have to do these experiments underground} \\
		\em \tiny Provide Feedback & \em \tiny Agreeing Statement & \em \tiny Main Topic &  \\
		\addlinespace
		\bottomrule
	\vspace{1.88cm}
	\end{tabular}
	\hspace{5pt}
	\begin{tabular}{lp{1.5cm}p{2cm}p{1.5cm}}
		\toprule
		{\bf Dialogue \#2} & {\bf Rating:} 2 &&\\
		\midrule
		\multicolumn{4}{p{7cm}}{{\bf Explainee:} how we extract meaning from the language we read? Do we link words to pictures in our mind?} \\
		\em \tiny Request Explanation & \em \tiny How Question & \em \tiny Main Topic & \\
		\addlinespace
		
		\multicolumn{4}{p{7cm}}{{\bf Explainer:} Your brain stores knowledge more as abstract concepts. The word links to that concept, and the image links to that concept. See a cat and your brain identifies it as a cat then gives you the thought "that's a cat". See the word cat and your brain identifies the meaning of a cat and then gives you the image of a cat in your imagination. These concepts are stored without the use of either word or image, but are linked to the separate storages of the words and images, so those systems usually fire together. This is also how you can know what you want to say but not quite recall the word for it - your brain has accessed the abstract concept, but has misplaced the link between it and the word for it.} \\
		\em \tiny Provide Explanation & \em \tiny Informing Statement & \em \tiny Main Topic & \\
		\addlinespace
		
		\multicolumn{4}{p{7cm}}{{\bf Explainee:} surely the image is the concept? I am sure for certain, attributes of an object are linked to the image rather than its name. Like you can know what to do with a pair of scissors even if you don't know their name in english.} \\
		\em \tiny Provide Explanation & \em \tiny Confirming Answer& \em \tiny Subtopic & \\
		\addlinespace
		
		\multicolumn{4}{p{7cm}}{{\bf Explainer:} Nope. In fact, there are people who have literally no ability to form a mental image at all, and yet who still have normal ability to understand what things mean.} \\
		\em \tiny Provide Explanation & \em \tiny Disagreeing Statement & \em \tiny Suptopic &  \\
			\addlinespace
			
		\multicolumn{4}{p{7cm}}{{\bf Explainee:} how so, please explain?} \\
		\em \tiny Request Explanation & \em \tiny How Question & \em \tiny Other&  \\
		\addlinespace
		
		\multicolumn{4}{p{7cm}}{{\bf Explainer:} It's not yet known exactly what causes this, we only know that people like this exist.} \\
		\em \tiny Provide Feedback & \em \tiny Disconfirming Answer & \em \tiny Other&  \\
		\addlinespace
		\bottomrule
		\end{tabular}
	\caption{Two example dialogues from our dataset that were rated with a high score of 4 (\#1) and a low score of 2 (\#2) by the annotators.}
	\label{table:example-dialogues}
\end{table*}

\begin{table}[t]%
	\centering
	\small
	\setlength{\tabcolsep}{1.5pt}%
	\renewcommand{\arraystretch}{1}
	\begin{tabular}{lrrrrrr}
		\toprule
		& & \multicolumn{5}{c}{\bf Score Distribution} \\
		\cmidrule(lr){3-7}
		{\bf Topic Relation} & Freq. & 1 & 2 & 3 & 4 & 5 \\
		\midrule
		(T01) Main Topic & 1747 & 26\% & 14\% & 18\% & 16\% & \bf 26\% \\
		(T04) No topic & 623 & \bf 47\% & 11\% & 12\% & 11\% & 19\% \\
		(T02) Subtopic & 611 & 25\% & 18\% & 31\% & 16\% & 10\% \\
		(T03) Related Topic & 476 & 25\% & 12\% & 24\% & 19\% & 20\% \\
		\bottomrule
	\end{tabular}
	\caption{The frequency of topic relation labels in our dataset broken  into each of the explanation quality levels [1-5]. Highlighted in bold values that distinguish the presence of these moves in high quality dialogues compared to low quality ones.}
	\label{table-topic-dist}
\end{table}

\end{document}